# Embedding Web-based Statistical Translation Models in Cross-Language Information Retrieval

Wessel Kraaij*
TNO TPD

Jian-Yun Nie†
Université de Montréal

Michel Simard‡
Université de Montréal

*Although more and more language pairs are covered by machine translation services, there are still many pairs that lack translation resources. Cross-language information retrieval (CLIR) is an application which needs translation functionality of a relatively low level of sophistication since current models for information retrieval (IR) are still based on a bag-of-words. The Web provides a vast resource for the automatic construction of parallel corpora which can be used to train statistical translation models automatically. The resulting translation models can be embedded in several ways in a retrieval model. In this paper, we will investigate the problem of automatically mining parallel texts from the Web and different ways of integrating the translation models within the retrieval process. Our experiments on standard test collections for CLIR show that the Web-based translation models can surpass commercial MT systems in CLIR tasks. These results open the perspective of constructing a fully automatic query translation device for CLIR at a very low cost.*

## 1 Introduction

Finding relevant information in any language on the increasingly multilingual World Wide Web poses a real challenge for current information retrieval (IR) systems. We will argue that the Web itself can be used as a translation resource in order to build effective cross-language IR systems.

### 1.1 Information Retrieval and Cross-Language Information Retrieval

The goal of IR is to find relevant documents from a large collection of documents, or from the World Wide Web. To do this, the user typically formulates a query, often in free text, to describe the information need. The IR system then compares the query with each document in order to evaluate its similarity (or probability of relevance) to the query. The retrieval result is a list of documents presented in decreasing order of similarity. The key problem in IR is that of effectiveness, i.e. how good an IR system is at retrieving the relevant documents and discarding the irrelevant ones.

Due to the information explosion on the Web, people are more in need of effective IR systems than ever before. The search engines on the Web are IR systems that have been created to answer this need. People are able to identify quickly the documents containing the same keywords as the query. However, the existing search engines only provide monolingual IR i.e. retrieve documents in the same language as the query. To be more precise: search engines usually do not consider the language of the keywords when the keywords of a query are matched against those of the documents. Identical keywords are matched whatever their languages are. For example, the English word "son" can match the French word "son" (his or her). Current search engines do not provide the functionality for Cross-Language IR (CLIR), i.e. the ability to retrieve relevant documents written in languages different from that of the query (without this latter being translated manually).

---

* TNO TPD, PO BOX 155, 2600 AD Delft, The Netherlands
† DIRO, Université de Montréal, CP. 6128, succ. Centre-ville, Montreal, Qc. H3C 3J7 Canada
‡ DIRO, Université de Montréal, CP. 6128, succ. Centre-ville, Montreal, Qc. H3C 3J7 Canada

With the growth of the Web, more and more documents on the Web are written in languages other than English, and many Internet users are non-native English speakers. For many users, the language barrier represents a serious problem. Although many users can read and understand rudimentary English, they feel uncomfortable in formulating queries in English, either because of their limited vocabulary in English, or because of the possible mis-usage of English words. For example, a Chinese user may use "economic" instead of "cheap/economical/inexpensive" in a query because these words have a similar translation in Chinese. An automatic query translation tool would be very helpful to these users. On the other hand, even if a user masters several languages, it is still a burden for him/her to formulate several queries in different languages. A query translation tool would also allow such a user to retrieve relevant documents in all the languages of interest with only one query. Even for the users with no understanding of a foreign language, a CLIR system may still be useful. For example, someone monitoring a competitor's developments on similar products may be interested in retrieving documents describing the possible products, even if he/she does not understand them. Machine translation systems may be used to gist the contents of the documents. For all these types of users, CLIR would represent a useful tool.

### 1.2 Possible approaches to CLIR

From an implementation point of view, the only difference between CLIR and the classical IR task is that the query language differs from the document language. It is obvious that to perform this task in an effective way, some form of translation is required. One might conjecture that a combination of two existing fields - IR and Machine Translation (MT) - would be satisfactory. One could simply translate the query by means of an MT system, obviating the need for a special CLIR system.

This approach, although feasible, is not the only possible approach, nor necessarily the best one. MT systems try to translate text into a well readable form, which is governed by morphological, syntactic and semantic constraints. However, current IR models are based on bag-of-words models. They are insensitive to word order and to the syntactic structure of the query. For example, the query "computer science" would usually produce the same retrieval results as "science computer". The complex process used in MT for producing a grammatical translation is not fully exploited by current IR models. This means that a simpler translation approach may suffice.

On the other hand, MT systems are far from perfect. Wrong translations are often produced. For example, Systran[1] translates the word "drug" as "drogue" (illegal substance) in French for both "drug traffic" and "drug administration office". Such a translation error will have a substantial negative impact on CLIR effectiveness. So even if MT systems are used as translation devices, they may need to be complemented by other more robust translation tools. In the current study, we will use statistical translation models as such a complementary tool.

---

1 We used the free translation service provided at http://babelfish.altavista.com/ in October 2002.

It is a well known fact that queries submitted to IR systems or search engines are often very short. In particular, the average length of queries submitted to the search engines on the Web is about 2 words (Jansen et al., 2001). Such short queries are generally insufficient to describe the information need in a precise and unambiguous way. Many important words are missing. For example, a user might formulate the query "Internet connection" in order to retrieve documents about "computer networks", "Internet Service Provider" or "proxy". However, under the current bag-of-words approach, the relevant documents containing the latter terms are unlikely to be retrieved. To solve this problem, a common approach used in IR is query expansion, which tries to add synonyms or related words to the original query, making the expanded query a more exhaustive description of the information need. The words added to the query during query expansion do not need to be strict synonyms. However, they have to be related to the information need to some degree. Ideally, the degree of the relatedness should be weighted so that a strongly related word is weighted higher than a less related one.

MT systems act in a way opposite to the query expansion process: only one translation is generally selected to express a meaning.[2] In doing so, MT systems in fact restrict the possible query expansion effect during the translation process. We believe that CLIR can benefit from a query translation which contains multiple translations for the same meaning. In this regard, the tests carried out by Kwok (Kwok, 1999) with a commercial MT system for Chinese-English CLIR are quite interesting. His experiments show that it is much better to use the intermediate translation data produced by the MT system than the final translation itself. The intermediate data contain, among other things, all the possible translation words. This work clearly demonstrates that using an MT system as a black box is not the best choice for query translation in CLIR. However, few MT systems allow one to access the intermediate stages of the translation.

Apart from the MT approach, query translation can also be done by using a machine-readable bilingual dictionary or by exploiting a set of parallel texts (texts with their translations). High quality bilingual dictionaries are expensive, but there are many free on-line translation dictionaries available on the Web, which can be used for query translation. This approach has been applied in several studies (e.g. (Hull and Grefenstette, 1996; Hiemstra and Kraaij, 1999)). However, free bilingual dictionaries often suffer from a poor coverage of the vocabulary, and from the problem of translation ambiguity because usually no information is provided to allow for disambiguation. In several previous studies (e.g. (Nie et al., 1999)), it has been shown that using a translation dictionary alone would produce much lower effectiveness than an MT system. However, a dictionary may be complemented by a statistical language model (Gao et al., 2001; Xu, Weischedel, and Nguyen, 2001). This has produced much better results than when the dictionary is used alone.

In this paper, the use of a bilingual dictionary is not our focus. We will concentrate on the third alternative for query translation: the approach based on parallel texts. Parallel texts are *texts accompanied by their translation in one or several other languages* (Véronis, 2000). They contain valuable translation examples for both human and machine translation. A number of studies in recent years (e.g. (Nie et al., 1999; Franz et al., 2001; Sheridan, Ballerini, and Schäuble, 1998; Yang et al., 1998)) have explored the possibility of using parallel texts for query translation in CLIR. One potential advantage of such an approach is that it provides multiple translations for the same meaning. The translation of a query would then contain not only true translation words, but also related words.

---

[2] Although there is no inherent obstacle preventing MT systems from generating multiple translations, in practice, only one translation is produced.

This is the query expansion effect that we want to produce in IR. Our experimental results have confirmed that this approach can be very competitive to the MT approach and yield much better results than an simple dictionary-based approach, while keeping the development cost low.

However, the one major obstacle to the use of parallel texts for CLIR is the unavailability of large parallel corpora for many language pairs. Hence, our first goal in this research was to develop an automatic mining system that collects parallel pages on the Web. The collected parallel Web pages are used to train statistical translation models (TM) that are then applied to query translation. Such an approach offers the advantage of enabling us to build a CLIR system for a new language pair, without waiting for the release of an MT system for that language pair. The number of potential language pairs supported by Web based translation models is large if one includes transitive translation using English as a pivot language. English is often one of the languages of parallel Web pages, when translations of Web pages are available.

The main objectives of this paper are twofold: i) we will show that it is possible to obtain large parallel corpora from the Web automatically, which can form the basis for an effective CLIR system, ii) we will compare several ways to embed the translation models in the IR system in order to exploit the corpora for cross-language query expansion.

Our experiments will show that these translation tools can result in CLIR effectiveness comparable to that of an MT system. This in turn demonstrates the feasibility of exploiting the Web as a large parallel corpus for the purpose CLIR.

**1.3 The problems in query translation**
Now let us turn to the query translation problem. In previous studies on CLIR three problems have been identified for query translation (Grefenstette, 1998b): identifying possible translations, pruning unlikely translations, and weighting the translation words.

**Finding translations**  First of all, whatever the translation tool employed, it has to provide a good coverage of the source and target vocabularies. In a dictionary-based approach to CLIR, we will encounter the same problems that have been faced in MT research: phrases, collocations, idiom and domain specific terminology are often translated incorrectly. These classes require a sophisticated morphological analysis and furthermore the domain-specific terms challenge the lexical coverage of the transfer dictionaries. A second important class of words which can pose problems for CLIR is the class of proper names, particularly for news article retrieval. Named entities such as the names of persons or locations are frequently used in queries for news articles and their translation is not always trivial. Often, the more commonly used geographical names of countries or capitals have a different spelling in other languages (Milan/Milano/Milaan), or translations that are not related to the same morphological root (Germany/Allemagne/Duitsland). The names of organizations and their abbreviations are also a notorious problem, e.g. the United Nations can be referred to as UN , ONU, VN etc. (disregarding the problem of morphological normalization of abbreviations). When proper names have to be translated from languages in a different script like Cyrillic, Arabic or Chinese, this problem is even more acute. The process of defining the spelling in a language with a different script is called transliteration and is based on a phonemic representation of the named entity. Unfortunately different national "standards" are used for transliteration, e.g. the former Russian president's name in Latin script has been transliterated as Jeltsin, Eltsine, Yeltsin, Jelzin etc.

**Pruning translation alternatives** A word or a term often has multiple translations. Some of them are appropriate for the query and the others are not. An important question is how to keep the appropriate translations while eliminating the inappropriate ones. Because of the particularities of IR it might help to keep multiple translations that display small sense differences, as in query expansion. So it could be beneficial to keep all these related senses for the matching process, together with their probabilities.

**Weighting translation alternatives** Closely related to the previous point is the question of how to deal with translation alternatives. The weighting of words in documents and in the query is of crucial importance in IR. A word with a high weight will influence the retrieval results more than a low-weight word. In CLIR it is also important to assign appropriate weights to translation words. Pruning translations can be viewed as an extreme Boolean way of weighting translations. The intuition is that, just like in query expansion, it may well be beneficial to assign a higher weight to the "main" translation and a lower weight to related translations.

**1.4 Integration of query translation with retrieval**

The problem of "weighting of translation alternatives" identified by Grefenstette refers to the more general problem of designing an architecture for a CLIR-system where translation and document ranking are integrated in a way that maximizes retrieval effectiveness.

The MT approach clearly separates the translation from the retrieval: the query is first translated, and the translation result is subsequently submitted to an IR system as a new query. At the retrieval phase, one no longer knows how certain a translated word is with respect to the other translated words in the translated query. All the translation words hare treated as though they are totally certain. Indeed, an MT system is used as a black box. In this paper, we consider translation to be an integral part of the IR process, which has to be considered together with the retrieval step.

From a more theoretical point of view, CLIR is a whole process which is composed of: query translation, document indexing and document matching. The two first sub-processes try to transform the query and the documents into a comparable internal representation form. The third sub-process tries to compare the representations to evaluate the similarity. In previous studies on CLIR, the first sub-process is clearly separated from the last two, which are integrated in classical IR systems. An approach that considers all the three sub-processes together will have the advantage of better accounting for the uncertainty of translation during the retrieval. More analysis on this point is provided in (Nie, 2002). This paper follows the same direction. We will show in our experiments that an integrated approach can produce very good CLIR effectiveness.

An attractive framework to integrate translation and retrieval is the probabilistic framework, although estimating the translation probabilities is not always straightforward.

In summary, because CLIR does not necessarily require a unique translation of a text (as MT does), approaches other than fully automatic MT might provide interesting characteristics for CLIR that are complementary to MT approaches. This could result in a higher precision, [3] since an MT system might choose the wrong translation, and/or a higher recall, [4] since multiple translations are accommodated, which could retrieve

---

3 Precision is defined as the proportion of relevant documents among all the retrieved documents.
4 Recall is the proportion of relevant documents retrieved among all the relevant documents in the documents via the related terminology.

In this paper we will investigate the effectiveness of CLIR systems based on probabilistic translation models trained on parallel texts mined from the Web. Globally, our approach to the CLIR problem can be viewed informally as "cross-lingual sense matching". Both query and documents are modeled as a distribution over semantic concepts, which in reality is approximated by a distribution over words. The challenge for CLIR is to measure to what extent these concept (or word sense) are related. From this point of view, our approach is similar in principle to that using Latent Semantic Analysis (LSI) (Dumais et al., 1997), which also tries to create semantic similarity between documents, queries and terms by transposing them into a new vector space. An alternative way of integrating translation and IR is to create so-called "structured queries", where translations are modeled as synonyms (Pirkola, 1998). Since this approach is simple and effective, we will use it as one of the reference systems in our experiments.

In this paper, this general approach will be implemented in several different ways, each fully embedded in the retrieval models. A series of experiments on CLIR will be conducted in order to evaluate these models. The results clearly show that Web-based translation models are as effective as (and sometimes more effective than) off-the-shelve commercial MT systems.

The remainder of the paper is organized as follows: Section 2 discusses the procedure we used to construct parallel corpora from the Web, and Section 3 describes the procedure used to train the translation models. Section 4 describes the probabilistic IR model that we employed and various ways of embedding translation into a retrieval model. Section 5 presents our experimental results. The article ends with a discussion and conclusion section.

**2 PTMINER**

It has been shown that by using a large parallel corpus, one can produce CLIR effectiveness close to the one obtained with an MT system (Nie et al., 1999). In previous studies, parallel texts have been exploited in several ways:

**With a pseudo feedback approach** In (Yang et al., 1998) parallel texts are used as follows. A given query in the source language is first used to retrieve a subset of texts from the parallel corpus. The corresponding subset in the target language is considered as providing a description of the query in the target language. From this subset of documents, a set of weighted words is extracted, and it is used as the query "translation".

**Capturing global cross language term associations** A more advanced and theoretically better motivated approach is to index concatenated parallel document in the dual space of the generalized vector space model (GVSM), where terms are indexed by documents (Yang et al., 1998). An approach related to GVSM is to build a so-called similarity thesaurus on the parallel or comparable corpus. A similarity thesaurus is an information structure (also based on the "dual space" of indexing terms by documents) which computes associated terms on the basis of global associations between terms as measured by term co-occurrence on the document level (Sheridan, Ballerini, and Schaüble, 1998). Recently, the idea of using the dual space of parallel documents for cross-lingual query expansion has been recast in a language modeling framework (Lavrenko, Choquette, and

---

collection.

Croft, 2002).

**Transposition to a language independent semantic space** The concatenated documents can also be transposed in a language-independent space by applying latent semantic indexing (Dumais et al., 1997; Yang et al., 1998). The disadvantage of this approach is that the concepts in this space are hard to interpret and that LSI is computationally demanding. It is currently not feasible to do this on a Web scale.

**To train a statistical translation model** This approach has been explored in e.g. (Nie et al., 1999; Franz et al., 2001). Statistical translation models (usually IBM model 1) are trained on the parallel corpus. The models are used in a straightforward way: the source query is submitted to the translation model, which proposes a set of translation equivalents, together with their probability. The latter are used as a new query for the retrieval process. In comparison with the first approach (Nie et al., 1999), this second approach (Franz et al., 2001) is based on a better founded theoretical framework: the OKAPI probabilistic IR model (Robertson and Walker, 1994). The present study is based on a different probabilistic IR model, based on statistical language models (Hiemstra, 2001; Xu, Weischedel, and Nguyen, 2001). This IR model facilitates a tighter integration of translation and retrieval. An important difference with the approaches based on document alignment discussed under the previous heading is that translation models perform alignment at a much more refined level. Consequently, the alignments can be used to estimate translation relations in a reliable way. On the other hand, the advantage of the CLIR approaches that just rely on alignment at the document level is that they can also handle comparable corpora, i.e. documents that discuss the same topic but are not necessarily translations of each other (Laffling, 1992).

Most previous work on parallel texts has been conducted on a few manually constructed parallel corpora, notably the Canadian Hansard corpus. This corpus [5] contains many years' debates in the Canadian parliament in both English and French, amounting to several dozens of millions words in each language. The European parliament documents represent another large parallel corpus in several European languages. However, its availability is much more restricted than the Canadian Hansard. For Chinese and English, the Hong Kong government publishes official documents in both Chinese and English. They form a Chinese-English parallel corpus; but again, its size is much smaller than the Canadian Hansard. For many other languages, no large parallel corpora are available for the training of statistical models.

LDC has tried to collect additional parallel corpora, resorting at times to manual collection (Ma, 1999). Several other research groups (for example, the RALI lab at Université de Montréal) also try to acquire manually constructed parallel corpora. However, manual collection of large corpora is a tedious task which is time- and resource-consuming. On the other hand, we observe that the increasing usage of different languages on the Web results in more and more bilingual and multilingual sites. Many Web pages are translated into different languages. The Web contains a large number of parallel Web pages for many languages (usually with English). If these can be extracted automatically, then this would help solve the problem of parallel corpora to some extent. PTMiner (for Parallel Text Miner) was built precisely for this purpose.

---

5 LDC provides a version containing texts from the mid-1970's through 1988, see `http://www.ldc.upenn.edu/`.

Of course, an automatic mining program is unable to understand the texts, and hence to judge if they are parallel in a totally reliable way. However, CLIR is quite error-tolerant. As we will show, a noisy parallel corpus can still be very useful for CLIR.

**2.1 General principles of automatic mining**

Parallel Web pages usually are not published in isolation, they are often connected in some way. For example, Resnik (Resnik, 1998) observed that some parallel Web pages are often referenced in the same parent index Web page. In addition, the anchor text of such links usually identifies the language. For example, if a Web page "index.html" contains links to both English and French versions of the referenced page, and the anchor texts of the links are respectively "English version" and "French version", then the referenced pages are probably parallel pages in English and French. To locate such pages, Resnik first sends a query of the following form to the Web search engine Alta Vista which returns the parent indexing pages:

```
anchor: English AND anchor: French
```

Then the referenced pages in both languages are retrieved and considered to be parallel. Applying this method, Resnik was able to mine 2491 pairs of English-French Web pages. Other researchers have adapted his system to mine 3376 pairs of English-Chinese pages and 59 pairs of English-Basque pages.

We observe, however, that only a small portion of parallel Web sites are organized in this way. Many other parallel pages cannot be found with this method. Our mining system uses different criteria; and we also incorporate an exploration process (i.e. a host crawler) in order to discover more Web pages that have not been indexed by the existing search engines.

The mining process in PTMiner is divided into two main steps: identification of candidate parallel pages, and verification of their parallelism. The overall process is organized into the following steps:

**Determining candidate sites** This step tries to identify the Web sites that may contain parallel pages. In our approach, we adopt a simple definition of Web site: it is a host corresponding to a distinct DNS (Domain Name System) address (e.g. www.altavista.com and geocities.yahoo.com).

**File name fetching** Identify a set of Web pages on each Web site that are indexed by search engines.

**Host crawling** Use the URLs collected in the previous step as seeds to further crawl each candidate site for more URLs.

**Pair scanning by names** Construct pairs of Web pages on the basis of pattern matching between URLs (e.g. "index.html" vs. "index_f.html").

**Text filtering** The candidate parallel pages are further filtered according to several criteria that operate on their contents.

In the following sub-sections, we describe each of these steps in more detail.

**2.2 Identification of Candidate Web Sites**

In addition to the organization of parallel Web pages used by Resnik, another common characteristic of parallel Web pages is that they cross-reference each other. For example, an English Web page may contain a pointer to the French version, and vice versa, and the anchor text of these pointers usually indicates the language of the other page. This phenomenon is common because such an anchor text shows the reader that a version in another language is available.

In considering both ways of organizing parallel Web pages, we see that a common feature is the existence of a link with an anchor text identifying a language. This is the criterion we use in PTMiner to detect candidate Web sites: the presence of at least one Web page containing such a link. They are identified via requests sent to a search engine e.g. AltaVista or Google. For example, the following request asks for pages in English that contain a link with one of the required anchor texts.

```
anchor: French version, in French, en Français, ...
language: English
```

The hosts extracted from the answers are considered to be candidate sites.

### 2.3 File Name Fetching

It is assumed that parallel pages are stored on the same Web site. This is not always true, but this assumption allows us to minimize the exploration of the Web and to avoid considering many unlikely candidates.

To search for parallel pairs from each candidate site, PTMiner first asks the search engine for all the Web pages from this site that they have indexed, via a request of the following form:

```
host: <hostname>
```

However, the results of this step may not be exhaustive because:

- search engines typically do not index all the Web pages of a site;

- most search engines allow users to retrieve a limited number of documents (e.g. 1,000 in AltaVista).

Therefore, we continue our search with a host crawler, which uses the Web pages found by the search engines as seeds.

### 2.4 Host Crawling

A host crawler is slightly different from a Web crawler or a robot in that a host crawler can only exploit one Web site at a time. A breadth-first crawling algorithm is used in this step. The principle is that if a retrieved Web page contains a link to an unexplored document on the same site, this document is added to the list of pages to be explored later. This crawling step allows us to obtain more Web pages from the candidate sites.

### 2.5 Pair Scanning by names

Once a large set of URLs has been identified, the next task is to determine parallel pairs from them. In our experience, we observed that many parallel Web pages have very similar file names. For example, an English Web page with the file name `index.html` often corresponds to a French translation with a file name such as `index_f.html`. The only difference between the two file names is a segment that identifies the language of the file. This similarity in file names is by no means an accident. In fact, this is a common way for Webmasters to keep track of a large number of documents in different versions.

This same observation also applies to URL paths. For example, the following two URLs are also similar in name:

`http://www.asite.ca/en/afile.html` vs. `http://www.asite.ca/fr/afile.html`.

To determine the name similarity of URLs, we define lists of prefixes and suffixes for both the source and the target languages. For example:

```
English Prefix = {(empty char), e, en, english, e_, en_, english_, ...}
```

Once a possible source language prefix is identified in an URL, it is replaced by a prefix in the target language, and we then test if this URL is found on the Web site.

**2.6 Filtering by contents**
The identified file pairs are further verified in terms of their contents. In PTMiner, the following criteria are used: file length, HTML structure and language verification.

**2.6.1 Text Length** The lengths of a pair of parallel pages are usually comparable to the typical length ratio of the two languages (especially when the text is long enough). Hence, a simple verification is to compare the lengths of the two files. As many Web documents are quite short, we tolerate some difference (up to 40% with the typical ratio).

**2.6.2 HTML Structure** Parallel Web pages are usually designed to have a similar layout. This often means that the two parallel pages have similar HTML structures. However, the HTML structures of parallel pages may also be quite different. Although pages look similar, they may still have different HTML markups. Therefore, certain flexibility is also allowed in this step.

In our approach, we first determine a set of meaningful HTML tags that affect the appearance of the page, and extract them from both files (e.g. `<p>` and `<H1>`, but not `<meta>` and `<font>`). A "diff"-style comparison will reveal how different the two sequences of tags are. A threshold is set to filter out the pairs that are not similar enough in HTML structure.

At this stage, non-textual parts of the pages are also removed. If a page does not contain enough text, it is also discarded.

**2.6.3 Language and Character Set** When we query search engines for documents in one specific language, the returned documents may be actually in a different language. This problem is particularly serious for Asian languages. When we ask for Chinese Web pages we often obtain Korean Web pages. This is because the language of the documents has not been identified accurately. Another more important factor that makes it necessary to use a language detector is that during host crawling and pair scanning, no verification is done on languages. All files with the _en suffix in their name are assumed to be an English page, which may be false.

To filter out the files not in the required languages, the SILC system (Isabelle, Simard, and Plamondon, 1998) is used. SILC employs n-gram statistical language models to determine the most probable language and encoding schema for a text. It has been trained on several large corpora for each language. The accuracy of the system is very high. When a text contains at least 50 characters, its accuracy is almost perfect. SILC can filter out a set of file pairs that are not in the required languages.

Our utilization of HTML structure to determine the parallelism of two pages is similar to that of (Resnik, 1998). Resnik also exploits an additional criterion similar to length-based sentence alignment in order to determine if the segments in the corresponding HTML structures have similar lengths. In the current PTMiner, this criterion is not incorporated. However, we have included the sentence-alignment criterion as a later filtering step in (Nie and Cai, 2001): If a pair of texts cannot be aligned reasonably well, then that pair is removed. This technique is shown to bring a large improvement for the English-Chinese corpus. A similar approach could also be envisioned for the corpora of European languages, but in the present study, it is not used.

**2.7 Mining results**

PTMiner uses heuristics that are mostly language-independent. This allows us to adapt it easily for different language pairs by changing a few parameters (e.g. prefix and suffix lists of file name). It is surprising that so simple an approach is nevertheless very effective. We have been able to construct large parallel corpora from the Web for the following language pairs: English-French, English-Italian, English-German, English-Dutch and English-Chinese. The size of these corpora are shown in Table 2.

|            | EN-FR   | EN-GE   | EN-IT  | EN-NL | EN-CH  |
|------------|---------|---------|--------|-------|--------|
| # Pairs    | 18,807  | 10,200  | 8,504  | 24738 | 14,820 |
| Size (MB)  | 174/198 | 77/100  | 50/68  | n.a.  | 74/51  |
| # Words (M)| 6.7/7.1 | 1.8/1.8 | 1.2/1.3| n.a.  | 9.2/9.9|

**Table 1**
Automatically mined corpora

One question that may be raised is how accurate, or how parallel, the mining results are. Actually, it is very difficult to answer this question. We have not undertaken an extensive evaluation, but only performed a simple evaluation with a set of samples. For English-French, from 60 randomly selected candidate sites, AltaVista indexed about 8,000 pages in French. From these, the pair-scan step identified 4,000 pages with equivalents in English. This showed that the lower bound of recall of pair-scan is 50%. The equivalence was judged by an undergraduate student who participated in developing the preliminary version of PTMiner. The criterion used to judge the equivalence between pages was subjective, the general guideline being whether two pages describe the same contents, and whether they have similar structures. To evaluate precision, 164 pairs were randomly selected and manually checked. It turned out that 162 of them were truly parallel. This shows that the precision is close to 99%.

For an English-Chinese corpus, a similar test has been reported in (Chen and Nie, 2000). This evaluation was done by a graduate student working on PTMiner. Among 383 pairs randomly selected at the pair-scan step, 302 pairs were really parallel. The precision ratio is 79%, which is not as good as that of the English-French case. There are several reasons for this:

- Incorrect links. It may be that the page is out-dated but still indexed by the search engines. Such a pair will be eliminated in the content filtering step.

- Pages that are designed to be parallel, although the contents are not all translated yet. One of the versions may be a simplified version of the other. Some of these cases can also be filtered out in the content filtering step; but some will still remain.

- Pages that are valid parallel pairs yet consist mostly of graphics instead of text. They therefore cannot be used for the training of translation models.

- Pairs that are not parallel at all. Their filenames accidentally match the naming rules. For example, `.../et.html` vs. `.../etc.html`.

Related to the last fact, we also observed that the names of Chinese and English pages may be very different. For example, it is frequent practice to use the Pinyin translation as the name of a Chinese page of the corresponding English file name (e.g. 'fangwen.html' vs. 'visit.html'). Another way is to use numbers as the filenames. For example '1.html' would correspond to '2.html'. In this case, our pair-scan approach based on name similarity will fail to recognize the pair. overall, the naming of Chinese files is much more variable and flexible. Hence, there exist fewer evident heuristics than for the European languages allowing us to enlarge the coverage and improve the precision of pair-scanning.

Given the potentially large amount of wrong parallel pairs, a question naturally arises: Can such a noisy corpus actually help CLIR? We will examine this question in Section 4. In the next section we will briefly describe how statistical translation models are trained on parallel corpora. We will focus on the following languages: English, French and Italian. The resulting translation models will be evaluated in a CLIR task.

## 3 Building the Translation Models

The bilingual pairs of documents collected from the Web are used as training material for the statistical translation models that we exploit for CLIR. In practice, this material must be organized into a set of small pairs of corresponding segments (typically, sentences), each consisting of a sequence of word tokens. We start by presenting the details of this preparatory step and then discuss the actual construction of the translation models.

### 3.1 Preparing the Corpus
**3.1.1 Format Conversion, Text Segmentation and Sentence Alignment** The collection process described in the previous section provides us with a set of pairs of HTML files. The first step in preparing this material is to extract the textual data from the files, and organize this data into small, manageable chunks (sentences).

In doing so, we try to take advantage of the HTML markup. For instance, we know that `<P>` tags normally identify paragraphs, `<LI>` tags mark list items which can also often be interpreted as paragraphs, `<Hn>` tags are normally used to mark section headers, and may therefore be taken as sentences, and so on.

Unfortunately, a surprisingly large number of HTML files on the Web are badly formatted, which calls for much flexibility on the part of Web browsers. To help cope with this situation, we employ to a freely-distributed tool called `tidy` (Ragget, 1998), which attempts to clean-up HTML files, so as to make them XML-compliant. This clean-up process mostly consists in normalizing tag-names to the standard XHTML lower-case convention, wrapping tag attributes within double-quotes and, most importantly, adding missing tags so as to end up with documents with balancing opening- and closing-tags.

Once this clean-up is done, we can parse the files with a standard SGML parser (we use `nsgmls` (Clark, 2001)), and use the output to produce documents in the standard `cesAna` format. This format, proposed as part of the Corpus Encoding Standard (CES (Ide, Priest-Dorman, and Vèronis, 1995)) is an SGML format with provisions for annotating simple textual structures such as sections, paragraphs and sentences. In addition to the cues provided by the HTML tags, we employ a number of heuristics, as well as language-specific lists of common abbreviations and acronyms to locate sentence boundaries within paragraphs.

When, as sometimes happens, the `tidy` programs fails to make sense of its input on a particular file, we simply remove all SGML markup from the file, and treat the document as plain-text, which means that we must rely solely on our heuristics to locate paragraph and sentence boundaries.

Once the textual data has been extracted from pairs of documents and is neatly segmented into paragraphs and sentences, we can proceed with sentence alignment. This operation produces what we call *couples*, i.e. minimal-size pairs of corresponding segments between two documents. In the vast majority of cases, couples consist of a single pair of sentences which are translations of one another (what we call "1-to-1" couples). However, there are sometimes "larger" couples, as when a single sentence in one language translates to two or more sentences in the other language ("1-to-N" or "N-to-1"), or when sentences translate many to many ("N-to-M"). Conversely, there are also "smaller" couples, such as when a sentence from either one of the two texts does not appear in the translation ("0-to-1" or "1-to-0").

Our sentence alignments are carried out by a program called `sfial`, an improved implementation of the method described in (Simard, Foster, and Isabelle, 1992). For a given pair of documents, this program uses dynamic programming to compute the alignment that globally maximizes a statistical-based scoring function. This function takes into account the statistical distribution of translation patterns ("1-to-1", "1-to-N", etc.), the relative sizes of the aligned text-segments, as well as the number of "cognate" words within couples, i.e. pairs of words with similar orthographies in the two languages (e.g. "statistical" in English v.s. "statistique" in French).

The data produced up to this point in the preparation process constitutes what we have called a **WAC** (*Web Aligned Corpus*).

**3.1.2 Tokenization, Lemmatization and Stop-words** Since our goal is to use translation models in an IR context, it seems natural to have both the translation models and the IR system operate on the same type of data. The basic indexing units of our IR systems are word stems. Stemming is an IR technique whereby morphologically related word-forms are reduced to a common form: a stem. Such a stem does not necessarily have to be a linguistic root form. The principal function of the stem is to serve as an index term in the vocabulary of index terms. Stemming is a form of conflation: equivalence classes of tokens help to reduce the variance in index terms. Most stemming algorithms fall into two categories: i) suffix strippers, and ii) full morphological normalization (sometimes referred to as "linguistic stemming" in the IR literature). Suffix strippers remove suffixes in an iterative fashion using rudimental morphological knowledge which is encoded in context sensitive patterns. The advantage of this type of algorithms (e.g. (Porter, 1980)) is their simplicity and efficiency, although this advantage applies principally to languages with a relatively simple morphology like English. A different way of generating conflation classes is to employ full morphological analysis. This process usually consists of two steps: first the texts are POS-tagged in order to eliminate each token's part-of-speech ambiguity and then word forms are reduced to their root form, a process which we refer to as lemmatization. More information about the relative utility of morphological normalization techniques in IR systems can be found in e.g. (Hull, 1996; Kraaij and Pohlmann, 1996; Braschler and Ripplinger, 2003).

Lemmatizing and removing stopwords from the training material is also beneficial for statistical translation modeling, helping to reduce the problem of data sparseness in the training set. Furthermore, function words and morpho-syntactic features typically arise from grammatical constraints intrinsic to a language, rather than as direct realizations of translated concepts. Therefore, we expect that removing them helps the translation model focus on meaning rather than form. In fact, it has been shown in (Chen and Nie, 2000) that the removal of stopwords from English-Chinese training material improves both the translation accuracy of the translation models and the effectiveness of CLIR. We expect a similar effect for European languages.

We also have to tokenize the texts, i.e. to identify individual word-forms. Because we are dealing with Romance languages, this step is fairly straightforward: [6] we essentially segment the text using blank spaces and punctuation. In addition, we rely on a small number of language-specific rules, for example to deal with elisions in French (*l'amour* → *l'* + *amour*) and Italian (*dell'arte* → *dell'* + *arte*), contractions in French (*au* → *à* + *le*), possessives in English (*Bob's* → *Bob* + *'s*), etc.

Once we have identified word-tokens, we can lemmatize or stem them. For Italian, we relied on a simple, freely-distributed stemmer from the Open Muscat project. [7] For French and English, we have access to more sophisticated tools that compute each token's lemma based on its part-of-speech (we use the HMM-based POS-tagger proposed in (Foster, 1991)) and extensive dictionaries with morphological information. As a final step, we remove stopwords.

Usually, 1-1 alignments are more reliable than other types of alignment. It is a common practice to use only this part for model training, and we do the same.

The following table provides some statistics on the processed corpora.

|                    | EN-FR       | EN-IT      |
|--------------------|-------------|------------|
| # 1-1 alignments   | 1018K       | 196K       |
| # tokens           | 6.7M/7.1M   | 1.2M/1.3M  |
| # unique stems     | 200K/173K   | 102K/87K   |

**Table 2**
Sentence-aligned corpora

### 3.2 Translation Models

In statistical translation modeling, we take the view that each possible target-language text is a potential translation for any given source-language text, but that some translations are more likely than others. In the terms of (Brown et al., 1990), a *noisy-channel translation model* is one that captures this state of affairs in a statistical distribution $P(T|S)$ where $S$ is a source-language text and $T$ is a target-language text. [8] With such a model, translating $S$ amounts to finding the target-language text $\hat{T}$ that maximizes $P(T|S)$.

Modeling $P(T|S)$ is, of course, complicated by the fact that there is an infinite number of possible source- and target-language texts, and so much of the work of the last 15 years or so in statistical machine translation has been aimed at finding ways to overcome this complexity by making various simplifying assumptions. Typically, $P(T|S)$ is rewritten as

$$P(T|S) = \frac{P(T)P(S|T)}{P(S)}$$

following Bayes' law. This decomposition of $P(T|S)$ is useful in two ways: first, it makes it possible to ignore $P(S)$ when searching for $\hat{T}$; second, it allows us to concentrate our efforts on the lexical aspects of $P(S|T)$, leaving it to $P(T)$ (the "target-language model") to take care of syntactic and other language-specific aspects.

In one of the simplest and earliest statistical translation models, IBM's *Model 1*, it is assumed that $P(S|T)$ can be approximated by a computation that uses only "lexical"

---

6 The processing on Chinese is described in (Chen and Nie, 2000).
7 Currently distributed by OMSEEK:
  http://cvs.sourceforge.net/cgi-bin/viewcvs.cgi/omseek/om/languages/
8 The model is referred to as *noisy-channel* because it takes the view that $S$ is the result of some input signal $T$ being corrupted while passing through a noisy channel. In this context, the goal is to recovered the initial input, given the corrupted output probabilities $P(s|t)$ over source- and target-language words $s$ and $t$. In other words, this model completely disregards the order in which the individual words of $S$ and $T$ appear.

While this model is known to be too weak for general translation, it appears that it can be quite useful for an application such as CLIR because many IR systems also disregard word-order, viewing documents and queries as unordered "bags of words".

The $P(s|t)$ distribution is estimated from a corpus of aligned sentences like the one we have produced from our Web-mined collection of bilingual documents, using the EM algorithm (*Expectation Maximization* (Baum, 1972)) to find the parameters that maximize the likelihood of the training set.

As in all machine-learning problems, especially those related to natural language, data sparseness is a critical issue in this process. Even with a large training corpus, many pairs of words $(s, t)$ occur at very low frequencies, and most never occur at all, making it impossible to obtain reliable estimates for the corresponding $P(s|t)$. Without adequate smoothing techniques, low-frequency events can have disastrous effects on the global behavior of the model; and unfortunately, in natural languages, low-frequency events are the norm rather than the exception.

The goal of translation in CLIR is different from that in general language processing. In the latter case it is important to enable a model to handle low-frequency words and unknown words. For CLIR the coverage of low-frequency words or unknown words by the model is less problematic. Even if such a word is translated incorrectly, the global IR effectiveness will often not be significantly affected because these words likely do not appear often in the document collection to be searched or other terms in the query could compensate for this gap. Most IR algorithms are based on a term weighting function which favors terms that occur frequently in a document but occur infrequently in the document collection. This means that the best index terms have a medium frequency (Salton and McGill, 1983). Stop-words and (near) hapaxes are less important for IR therefore, limited coverage of very infrequent words in a translation model is not critical for the performance of a CLIR system. Proper nouns are special cases of unknown words. When they appear in a query they usually denote an important part of the user's intention. However, we can adopt a special approach to cope with these unknown words in CLIR without integrating them as the generalized case in the model. For example, one can simply retain all the unknown words in the query translation. This approach works well for most cases in European languages. We have previously shown that a fuzzy matching approach based on n-grams offers an effective means of overcoming small spelling variations in proper noun spelling (Kraaij, Pohlmann, and Hiemstra, 2000).

The model pruning techniques developed in computational linguistics are also useful for the models used in CLIR. The beneficial effect is that unreliable (or low probability) translations can be removed. In Section 4, model smoothing will be motivated from a more theoretical point of view. Here, let us first outline the two variations we used to prune the models.

The first one is simple, yet effective in our application: we consider unreliable all parameters (translation probabilities) whose value falls below some pre-set threshold (in practice, $0.1$ works well). These parameters are simply discarded from the model. The remaining parameters are then re-normalized so that all marginal distributions sum to 1.

Another pruning technique is based on the relative contribution to the entropy of the model. We retain the $N$ most reliable parameters (in practice N=100K works well). The reliability of a parameter is measured with regard to its contribution to the model's entropy (Foster, 2000). In other words, we discard the parameters that least affect the overall probability of the training set. The remaining parameters are then re-normalized so that all marginal distributions sum to 1.

Of course, as a result of this, most pairs of words $(s, t)$ are unknown to the translation model (translation probability equals zero). As previously discussed, however, this will not have a disastrous effect on CLIR; on the contrary, some positive effect can be expected as long as there is at least one translation for each source term.

One important characteristic of these noisy-channel models is that they are "directional". Depending on the intended use, it must be determined beforehand which language is the source and which the target for each pair of languages. Although "reverse" parameters can theoretically be obtained from the model through Bayes' rule, it is often more practical to train two separate models if both directions are needed. This topic is also discussed in the next section.

## 4 Embedding translation into the IR model

When CLIR is considered simply as a combination of separate MT and IR components, the embedding of the two functions is not a problem. However, as we explained in the introduction, there are theoretical motivations for embedding translation into the retrieval model. Since translation models provide more than one translation, we will try to exploit this extra information, in order to enhance retrieval effectiveness. In section 4.1 we will first introduce a monolingual probabilistic IR model based on cross-entropy between a unigram language model for the query and one for the document. We discuss the relationship of this model with IR models based on generative language models. Subsequently, we show several ways to add translation to the model: one can either translate the query language model from the source language into the target language (i.e. the document language) before measuring the cross-entropy; or translate the document model from the target language into the source language and then measuring the cross entropy.

### 4.1 Monolingual IR based on unigram language models

Recently, a new approach to IR based on statistical language models has gained wide acceptance. The approach was developed independently by several groups (Ponte and Croft, 1998; Miller, Leek, and Schwartz, 1999; Hiemstra, 1998) and has yielded results comparable or better than the existing OKAPI probabilistic model on several IR standardized evaluation tasks. In comparison with the OKAPI model, the IR model based on generative language models has some important advantages: it contains fewer collection dependent tuning parameters and is easy to extend. For a more detailed discussion of the relationships between the classical (discriminative) probabilistic IR models and recent generative probabilistic IR models we refer the reader to (Kraaij and Spitters, 2003). Probably the most important idea in the language modeling approach to IR is that documents are scored on the probability that they generate the query, i.e. the problem is reversed, an idea which has successfully been applied in speech recognition. There are various reasons why this approach has proven fruitful, probably the most important being that documents contain much more data to estimate the parameters of a probabilistic model than do ad-hoc queries (Lafferty and Zhai, 2001b). For ad-hoc retrieval, one could describe the query formulation process as follows: a user has an ideal relevant document in mind and tries to describe it by mentioning some of the salient terms that he thinks occur in the document, interspersed with some query stop-phrasing like "Relevant documents mention..". For each document in the collection, we can compute the probability that the query is generated from a model representing that document. This generation process can serve as a coarse way of modeling the user's query formulation process. The query-likelihood given each document can directly be used as a document ranking function. Formula (1) shows the basic language model, where a query consist of a sequence of terms $T_1, T_2, \ldots, T_m$ which are sampled independently from a document unigram model for document $d_k$.

$$P(Q|D_k) = P(T_1, T_2, \ldots, T_m | D_k) \approx \prod_{j=1}^{m} P(T_m | M_{D_k}) \qquad (1)$$

In this formula $M_{D_k}$ denotes a language model of $D_k$. It is indeed an approximation of $D_k$. Now, if a query is more probable given a language model based on document $D_1$ than given a language model based on document $D_2$, we can then hypothesize that document $D_1$ is more likely to be relevant to the query than document $D_2$. Thus the probability of generating a certain query given a document-based language model can serve as a score to rank documents with respect to topical relevance. It is common practice to work with log probabilities, which has the advantage that products reduce to summations. We will therefore rewrite (1) in logarithmic form. Since terms might occur more than once in a query; we prefer to work with types $\tau_i$ instead of tokens $T_i$. So $c(Q, \tau_i)$ is the number of occurrences of $\tau_i$ in $Q$ (query term frequency), we will also omit the document subscript $k$ in the following presentation.

$$\log P(Q|D) = \sum_{i=1}^{n} c(Q, \tau_i) \log P(\tau_i | M_D) \qquad (2)$$

A second core technique from speech recognition that plays a vital role in language models for IR is smoothing. One obvious reason is to avoid assigning zero probabilities for terms that do not occur in a document because the term probabilities are estimated using maximum likelihood estimation. [9] If a single query term does not occur in a document, this would result in a zero probability of generating that query, which might not be desirable in many cases since documents discuss a certain topic using only a finite set of words. It is very well possible that a term which is highly relevant for a topic may not appear in a document, since it is a synonym. Longer documents will in most cases have a better coverage of relevant index terms (and consequently better probability estimates) than short documents, so one could let the level of smoothing depend on the length of the document (e.g. Dirichlet priors). A second reason for smoothing probability estimates of a generative model for queries is that queries consist of i) terms which have a high probability of occurrence in relevant documents and ii) terms which are merely used to formulate a proper query statement (e.g. "Documents discussing only X are not relevant."). A mixture of a document language model and a language model of typical query terminology (estimated on millions of queries) would probably give good results (in terms of a low perplexity).

We have opted for a simple approach that addresses both issues, namely applying a smoothing step based on linear interpolation with a background model estimated on a large document collection, since we do not have a collection of millions of queries.

$$\log P(Q|D) = \sum_{i=1}^{n} c(Q, \tau_i) \log((1-\lambda) P(\tau_i | M_D) + \lambda P(\tau_i | M_C)) \qquad (3)$$

---

[9] The fact that language models have to be smoothed seems to contradict the discussion in Section 3 where we stated that rare terms are not critical for IR effectiveness. This is not the case. Smoothing helps to make the distinction between absent important terms (mid-frequency terms) and absent non-important terms (high frequency terms). A document which misses important terms should be down-weighted in score more than a document which misses an unimportant term.

Here, $P(\tau_i|M_C)$ denotes the marginal probability of observing the term $\tau_i$, which can be estimated on a large background corpus, and $\lambda$ is the smoothing parameter. A common range for $\lambda$ is 0.5-0.7, which means that document models have to be smoothed quite heavily for optimal performance. We hypothesize that this is mainly due to the query modeling role of smoothing. Linear interpolation with a background model has been frequently used to smooth document models, e.g. (Miller, Leek, and Schwartz, 1999; Hiemstra, 1998). Recently other smoothing techniques (Dirichlet, absolute discounting) have also been evaluated. An initial attempt to account for the two needs for smoothing (sparse data problem, query modeling) with separate specialized smoothing functions yielded positive results (Zhai and Lafferty, 2002).

| symbol | explanation |
|---|---|
| $Q$ | Query has representation $Q = \{T_1, T_2, ..., T_n\}$ |
| $D$ | Query has representation $D = \{T_1, T_2, ..., T_n\}$ |
| $M_Q$ | Query language model |
| $M_D$ | Document Language model |
| $M_C$ | Background language model |
| $\tau_i$ | index term |
| $s_i$ | term in the source language |
| $t_i$ | term in the target language |
| $\lambda$ | smoothing parameter |
| $c(x)$ | counts of $x$ |

**Table 3**
Common symbols

We have tested the model corresponding to formula (3) in several different IR applications: monolingual information retrieval, filtering, topic detection and tracking (cf. (Allen, 2002) for a task description of the latter two tasks). For several of these applications (topic tracking, topic detection, collection fusion), it is important that scores be comparable across different queries (Spitters and Kraaij, 2001). For the basic model, this is not the case; so it has to be extended with score normalization. There are two important steps to do this. First of all, we would like to normalize across query specificity. The generative model will produce low scores for specific queries (since the average probability of occurrence is low) and higher scores for more general queries. This can be accomplished by modeling the IR task as a likelihood ratio (Ng, 2000). For each term in the query, the LLR (log likelihood ratio) model judges how much surprise there is to see this term given the document model in comparison with the background model.

$$\text{LLR}(Q|D) = \log \frac{P(Q|M_D)}{P(Q|M_C)} = \sum_{i=1}^{n} c(Q, \tau_i) \log \frac{((1-\lambda)P(\tau_i|M_D) + \lambda P(\tau_i|M_C))}{P(\tau_i|M_C)} \quad (4)$$

In (4), $P(Q|M_C)$ denotes the generative probability of the query given a language model estimated on a large background corpus $C$. Note that $P(Q|M_C)$ is a query dependent constant and does not affect document ranking. Actually, model (4) has a better justification than model (3), since it can be seen as a direct derivative of the log-odds of relevance if we assume uniform priors for document relevance:

$$\log \frac{P(R|D,Q)}{P(\bar{R}|D,Q)} = \log \frac{P(Q|R,D)}{P(Q|\bar{R},D)} + \log \frac{P(R|D)}{P(\bar{R}|D)} \approx \log \frac{P(Q|M_D)}{P(Q|M_C)} + K \quad (5)$$

In (5), $R$ refers to the event that a user likes a document, i.e. the document is relevant.

The scores of model (4) still depend on the query length, which can be easily normalized by dividing the scores by the query length ($\sum_i c(Q, \tau_i)$). This results in formula (6) for the normalized log likelihood ratio (NLLR) of the query:

$$\text{NLLR}(Q|D) = \sum_{i=1}^{n} \frac{c(Q, \tau_i)}{\sum_i c(Q, \tau_i)} \log \frac{((1-\lambda)P(\tau_i|M_D) + \lambda P(\tau_i|M_C))}{P(\tau_i|M_C)} \qquad (6)$$

A next step is to view the normalized query term counts $c(Q, \tau_i)/\sum_i c(Q, \tau_i)$ as maximum likelihood estimates of a probability distribution representing the query $P(\tau_i|M_Q)$. The NLLR formula can now be reinterpreted as a relationship between the two probability distributions $P(\tau|M_Q)$, $P(\tau|M_D)$ normalized by the the third distribution $P(\tau|M_C)$. The model measures how much better than the background model the document model can encode events from the query model; or in information theoretic terms, it can be interpreted as the difference between two cross-entropies:

$$\text{NLLR}(Q|D) = \sum_{i=1}^{n} P(\tau_i|Q) \log \frac{P(\tau_i|D_k)}{P(\tau_i|C)} = H(X|c) - H(X|d) \qquad (7)$$

$X$ is a random variable with the probability distribution $p(\tau_i) = p(\tau_i|M_Q)$ and $c$ and $d$ are probability mass functions representing the marginal distribution and the document model. Cross-entropy is a measure of our average surprise; so the better a document model 'fits' a query distribution, the higher the score will be. [10]

The representation of both the query and a document as samples from a distribution representing respectively the user's request and the document author's "mindset" has several advantages. Traditional IR techniques like query expansion and relevance feedback can be reinterpreted in an intuitive framework of probability distributions (Lafferty and Zhai, 2001a; Lavrenko and Croft, 2001). The framework also seems suitable for cross language retrieval. We only need to extend the model with a translation function, which relates the probability distribution in one language with the probability distribution function in another language. We will present several solutions for this extension in the next section.

The NLLR also has a Disadvantage: it is less easy to integrate prior information about relevance into the model (Kraaij, Westerveld, and Hiemstra, 2002), which can be done in a straightforward way in formula (1), by simple multiplication. CLIR is a special case of ad-hoc retrieval and usually a document length based prior can enhance results significantly. A remedy which has proven to be effective is linear interpolation of the NLLR score with a prior log-odds ratio $\log(P(R|D)/P(\neg R|D))$ (Kraaij, 2002). For reasons of clarity, we have chosen to not to include this technique in the experiments presented here.

In the following sections, we will describe several ways to extend the monolingual IR model with translation. The section headers contain the run tags that will be used in Section 5 to describe the experimental results.

**4.2 Estimating the query model in the target language (QT)**
In Section 4.1, we have seen that the basic retrieval model measures the cross-entropy between two language models: a language model of the query and a language model of the document. [11] Instead of translating a query before estimating a query model (the external approach), we propose to directly estimate the query model in the target language. We will do this by decomposing the problem into two components that are easier to estimate:

$$P(t_i|M_{Q_s}) = \sum_j^L P(s_j, t_i|M_{Q_s}) = \sum_j^L P(t_i|s_j, M_{Q_s})P(s_j|M_{Q_s}) \approx \sum_j^L P(t_i|s_j)P(s_j|M_{Q_s})$$
(8)

where $L$ is the size of the source vocabulary. Thus, $P(t_i|M_{Q_s})$ can be approximated by combining the translation model $P(t_i|s_j)$, which we can estimate on the parallel Web corpus, and the familiar $P(s_j|M_{Q_s})$ which can be estimated using relative frequencies.

---

[10] The NLLR can also be reformulated as a difference of two Kullback-Leibler divergences(Ng, 2000)
[11] We omit the normalization with the background model in the formula for presentation reasons.

This simplified model, from which we have dropped the dependency of $P(t_i|s_j)$ on $Q$, can be interpreted as a way of mapping the probability distribution function in the source language event space $P(s_j|M_{Q_s})$ onto the event space of the target language vocabulary. Since this probabilistic mapping function involves a summation over all possible translations, mapping the query model from the source language can be implemented as the matrix product of a vector representing the query probability distribution over source language terms with the translation matrix $P(t_i|s_j)$. [12] The result is a probability distribution function over the target language vocabulary.

Now we can substitute the query model $P(\tau_i|M_Q)$ in formula (7) with the target language query model in (8) and, after a similar substitution operation for $P(\tau_i|M_C)$, we arrive at CLIR-model QT:

$$\text{NLLR-QT}(Q_s|D_t) = \sum_{i=1}^n \sum_{j=1}^L P(t_i|s_j)P(s_j|M_{Q_s}) \log \frac{(1-\lambda)P(t_i|M_{D_t}) + \lambda P(t_i|M_{C_t})}{P(t_i|M_{C_t})}$$
(9)

**4.3 Estimating the document model in the source language (DT)**
Another way to embed translation into the IR model is to estimate the document model in the source language:

$$P(s_i|M_{D_t}) = \sum_j^N P(s_i, t_j|M_{D_t}) = \sum_j^N P(s_i|t_j, M_{D_t})P(t_j|M_{D_t}) \approx \sum_j^N P(s_i|t_j)P(t_j|M_{D_t})$$
(10)

where $N$ is the size of the target vocabulary. Obviously, we need a translation model in the reverse direction for this approach. Now we can substitute (10) for $P(\tau_i|M_D)$ in formula (7), yielding CLIR model DT:

$$\text{NLLR-DT}(Q_s|D_t) = \sum_{i=1}^n P(s_i|M_{Q_s}) \log \frac{\sum_{j=1}^N P(s_i|t_j)((1-\lambda)P(t_j|M_{D_t}) + \lambda P(t_j|M_{C_t}))}{\sum_{j=1}^N P(s_i|t_j)P(t_j|M_{C_t})}$$
(11)

It is important to realize that both the QT and DT models are based on context insensitive translation, since translation is added to the IR model after the independence assumption (1) has been made. Recently, a more complex CLIR model based on relaxed assumptions - context sensitive translation but term-independence based IR - has been proposed in (Federico and Bertoldi, 2002). In experiments on the CLEF test collections, the aforementioned model also proved to be more effective; however, it has the disadvantage of reducing efficiency due to a Viterbi search procedure.

---

12 For presentation reasons, we have replaced the variable $\tau$ used in Section 4.1 with $s$ and $t$ for a term in the source and target language respectively.

**4.4 Variant models and baselines**
In this section we will discuss several variant instantiations of the QT and DT model, which help us measure the importance of the number of translations (pruning) and the weighting of translation alternatives. We also present several baseline CLIR algorithms taken from the literature and discuss their relationship to the QT and DT models.

**4.4.1 External translation (MT, NAIVE)**  As we already argued in the introduction, the most simple solution to CLIR is to use an MT system to translate the query and use the translation as the basis for a monolingual search operation in the target language. This solution does not require any modification to the standard IR model as presented in Section 4.1. We will refer to this model as the *external* translation approach. The translated query is used to estimate a probability distribution for the query in the target language. Thus, the order of operations is: i) translate the query using an external tool; ii) estimate the parameters $P(t_i|M_{Q_t})$ of a language model based on this translated query.

In our experimental section below, we will list results with two different instantiations of the external translation approach: i) MT: query translation by Systran, which attempts to use high-level linguistic analysis, context-sensitive translation, extensive dictionaries etc. ii) NAIVE: naive replacement of each query term by its translations (not weighted). The latter approach is often implemented using bilingual word lists for CLIR. It is clear that this approach can be problematic for terms with many translations, since they would then get a higher relative importance. The NAIVE method is only included here to study the effect of the number of translations on the effectiveness of various models.

**4.4.2 Best match translation (QT-BM)**  In Section 3.2 we have already explained that there are different possible strategies to prune the translation model. An extreme pruning method is best-match, where only the best translation is kept. A best-match translation model for query model translation (QT-BM) could also be viewed as an instance of the external translation model, but one that uses a corpus-based disambiguation method. Each query term is translated by the most frequent translation in the Web corpus, disregarding the query context.

**4.4.3 Equal probabilities (QT-EQ)**  If we don't know the precise probability of each translation alternative for a given term, the best thing to do is to fall back on uniform translation probabilities. This situation arises, for example, if we only have standard bilingual dictionaries. We hypothesize that this approach will be more effective than NAIVE but less effective than QT.

**4.4.4 Synonym based translation (SYN)**  An alternative way to embed translation into the retrieval model is to view translation alternatives as synonyms. This is, of course, something of an idealization, yet there is certainly some truth to the approach when translations are looked up in a standard bilingual dictionary. Strictly speaking, when terms are pure synonyms, they can be substituted. Combining translation alternatives with the synonym operator of the INQUERY IR system (Broglio et al., 1995), which conflates terms on the fly, has been shown to be an effective way of improving the performance of dictionary-based CLIR systems (Pirkola, 1998). In our study of stemming algorithms (Kraaij and Pohlmann, 1996), we independently implemented the synonym operator in our system. This on-line conflation function replaces the members of the equivalence class by a class id, usually a morphological root form. We have used this function to test the effectiveness of a synonymy-based CLIR model in a language model IR setting.

The synonym operator for CLIR can be formalized as the following class equivalence model (assuming $n$ translations $t_j$ for term $s_i$ and $N$ unique terms in the target language):

$$P(\text{class}(s_i)|M_{D_t}) = \frac{\sum_j^n c(t_j, D_t)}{\sum_j^N c(t_j, D_t)} = \sum_j^N \delta(s_i, t_j) P(t_j|M_{D_t}) \quad (12)$$

where $P(\text{class}(s_i)|M_{D_t})$ is the probability that a member of the equivalence class of $s_i$ is generated by the language model $M_{D_t}$ and

$$\delta(s_i, t_j) = \begin{cases} 1 & \text{if } t_j \in \text{class}(s_i) \\ 0 & \text{if } t_j \notin \text{class}(s_i) \end{cases} \quad (13)$$

Here $c(t_j, D_t)$ is the term frequency (counts) of term $t_j$ in document $D_t$.

The synonym class function $\delta(s_i, t_j)$ can be interpreted as a special instantiation of the translation model $P(s_i|t_j)$ in (10), namely $P(s_i|t_j) = 1$ for all translations $t_j$ of $s_i$. Of course this does not yield a valid probability function since the translation probabilities for all translations $s_i$ of a certain $t_j$ do not sum to one, because the pseudo-synonym classes are not disjunct due to sense ambiguity. But the point is that the structure of a probabilistic version of the SYN model is similar to the DT model, namely one where all translations have a reverse translation probability $P(s_i|t_j)$ equal to one. This is obviously just an approximation of reality. We therefore expect that this model will be less effective than the QT and DT models. In our implementation of the SYN model, we formed equivalence classes by looking up all translations of a source term $s_i$ in the translation model $P(t_j|s_i)$. The translations receive weight 1 and are used as pseudo translation probabilities [13] in the model corresponding to formula (11).

### 4.5 Related work

In dictionary-based approaches, the number of translation alternatives is usually not as high as in (un-pruned) translation models, so these translations can be used in some form of query expansion (Hull and Grefenstette, 1996; Savoy, 2002). However, it is well known that most IR models break down when the number of translations is high.

To remedy this, researchers have tried to impose query structure, e.g. by collecting translation alternatives in an equivalence class (Pirkola, 1998), or via a quasi Boolean structure (Hull, 1997).

This idea of embedding a translation step into an IR model based on query likelihood was developed independently by several researchers (Hiemstra and de Jong, 1999; Kraaij, Pohlmann, and Hiemstra, 2000; Berger and Lafferty, 2000). Initially translation probabilities were estimated from machine-readable dictionaries, using simple heuristics (Hiemstra et al., 2001). Other researchers have successfully used models similar to DT, in combination with with translation models trained on parallel corpora, though not from the Web (McNamee and Mayfield, 2001; Xu, Weischedel, and Nguyen, 2001).

---

13 or maybe it is better to view them as mixing weights in this case

## 5 Experiments

We carried out several contrastive experiments to gain more insight into the relative effectiveness of the various CLIR models presented in Section 4.2 - 4.4. We will first outline our research questions, before describing the experiments in more detail.

### 5.1 Research Questions
The research questions we are hoping to answer are the following:

**i)** How do CLIR systems based on translation models perform w.r.t. reference systems (e.g. monolingual, MT )?

**ii)** Which manner of embedding a translation model is most effective for CLIR? How does a probabilistically motivated embedding compare with a synonym based embedding?

**iii)** Is there a query expansion effect and how can we exploit it?

**iv)** What is the relative importance of pruning versus weighting?

**v)** Which models are robust against noisy translations?

The first two questions concern the main goal of our experiments: What is the effectiveness of a probabilistic CLIR system in which translation models mined from the Web are an integral part of the model, compared to CLIR models in which translation is merely an external component? The remaining questions help to understand the relative importance of various design choices in our approach, such as pruning, translation model orientation etc.

### 5.2 Experimental conditions
We have defined a set of contrastive experiments in order to help us answer the above-mentioned research questions. These experiments seek to:

1. Compare the effectiveness of approaches incorporating a translation model produced from the Web versus a monolingual baseline and an off-the-shelf external query translation approach based on Systran (MT).

2. Compare the effectiveness of embedding query model translation (QT) and document model translation (DT).

3. Compare the effectiveness of using a set of all-weighted translations (QT) versus just the best translation (QT-BM).

4. Compare the effectiveness of weighted query model translation (QT) versus equally-weighted translations (QT-EQ) and non-weighted translations (NAIVE).

5. Compare the effectiveness of treating translations as synonyms (SYN) with weighted translations (QT) and equally-weighted translations (QT-EQ).

6. Compare different translation model pruning strategies: best $N$ parameters or thresholding probabilities.

Each strategy is represented by a run-tag, as shown in Table 4.

Table 5 illustrates the differences between the different translation methods. It lists, for several CLIR models, the French translations of the word "drug" taken from one of the test queries which talks about drug policy.

| run tag | short description | matching language | Section |
|---|---|---|---|
| MONO | monolingual run | | 4.1, 5.5 |
| MT | Systran external query translation | target | 4.4.1, 5.5 |
| NAIVE | equal probabilities | target | 4.4.1 |
| QT | translation of the query language model | target | 4.2 |
| DT | translation of the document language model | | 4.3 |
| QT-BM | best match, one translation per word | target | 4.4.2 |
| QT-EQ | equal probabilities | target | 4.4.3 |
| SYN | synonym run based on forward equal probabilities | source | 4.4.4 |

**Table 4**
Explanation of the run tags

The translations in Table 5 are provided by the translation models $P(e|f)$ and $P(f|e)$. The translation models have been pruned by discarding the translations with $P < 0.1$ and renormalizing the model (except for SYN), or by retaining the 100K best parameters of the translation model. The first pruning method (probability threshold) has a very different effect on the DT method in comparison with QT: the number of terms that translate into *drug* according to $P(e|f)$ is much larger than the translations of *drug* found in $P(f|e)$. There are several possible explanations for this: quite a few French terms, including the verb *droguer*, the compounds *pharmacorésistance, pharmacothérapie* etc., all translate into an English expression or compound involving the word *drug*. Since our translation model is quite simple, these compound-compound translations are not learned. [14] A second factor that might play a role is the greater verbosity of French text compared to their English equivalent (cf. Table 2). For the models which have been pruned using the 100K best parameters criterion, the differences between QT and DT are smaller. Both methods yield multiple translations, most of which seem related to *drug*; so there is a clear potential for improved recall due to the query expansion effect. Notice, however, that the expansion concerns both the medical and the narcotic senses of the word *drug*. We will see in the following section that the CLIR model is able to take advantage of this query expansion effect, even if the expansion set is noisy and not disambiguated.

**5.3 The CLEF test collection**
To achieve our objective, we carried out a series of experiments on a combination of the CLEF-2000, -2001 and -2002 test collections. [15] This joint test collection consists of documents in several languages (articles from major European newspapers from the year 1994), 140 *topics* describing different information needs (also in several languages) and their corresponding relevance judgments. We only used the English, Italian and French data for the CLIR experiments reported here. The main reason for this limitation was that the IR experiments and translation models were developed at two different sites equipped with different proprietary tools. We chose language pairs for which the lemmatization/stemming step for both the translation model training and indexing system were equivalent. A single test collection was created by merging the three topic-sets in order to increase the reliability of our results and sensitivity of significance tests. Each

---

14 A more extreme case is query C044 about the "tour de france". According to the $P(e|f) > 0.1$ translation model, there are 902 French words that translate into the "English" word *de*. This is mostly due to French proper names, which are left untranslated in the English parallel text
15 CLEF=Cross Language Evaluation Forum, `www.clef-campaign.org`

| run id | translation | translation model |
|---|---|---|
| MT | drogues | |
| QT | <drogue,0.55; medicament,0.45> | $P(f\|e) \leq 0.1$ |
| QT-EQ | <drogue,0.5; medicament,0.5> | |
| QT-BM | <drogue,1.0> | |
| SYN | <drogue,1.0; medicament,1.0> | |
| NAIVE | <drogue,1.0; medicament,1.0> | |
| DT | <antidrogue,1.0; drogue,1.0; droguer,1.0; drug,1.0; médicament,0.79; drugs,0.70; drogué 0.61; narcotrafiquants,0.57; relargage,0.53; pharmacovigilance,0.49; pharmacorésistance,0.47 médicamenteux,0.36; stéroïdiens,0.35, stupéfiant,0.34; assurance-médicaments,0.33; surdose 0.28; pharmacorésistants,0.28; pharmacodépendance,0.27 pharmacothérapie,0.25; alcoolisme,0.24; toxicomane,0.23; bounce,0.23; anticancéreux,0.22; anti-inflammatoire,0.17; selby,0.16; escherichia,0.14; homelessness,0.14; anti-drogues,0.14; antidiarrhéique,0.12; imodium,0.12; surprescription,0.10> | $P(e\|f) \leq 0.1$ |
| QT | <drogue,0.45; medicament,0.35; consommation, 0.06; relier, 0.03; consommer, 0.02; drug, 0.02; usage, 0.02; toxicomanie, 0.01; substance, 0.01; antidrogue, 0.01; utilisation, 0.01; lier, 0.01; thérapeutique, 0.01; actif, 0.01; pharmaceutique, 0.01> | $P(e\|f)$, 100K |
| DT | <reflexions, 1; antidrogue, 1; narcotrafiquants, 1; drug, 1; droguer, 0.87; drogue, 0.83; drugs, 0.81; médicament, 0.67; pharmacorésistance, 0.47; pharmacorésistants, 0.44; médicamenteux, 0.36; stupéfiant, 0.34; assurance-médicaments, 0.33; pharmacothérapie, 0.33; amphétamine, 0.18; toxicomane, 0.17; mémorandum, 0.10; toxicomanie, 0.08; architectural, 0.08; pharmacie, 0.07; pharmaceutique, 0.06; thérapeutique, 0.04; substance, 0.01> | $P(f\|e)$, 100K |

**Table 5**
Example translations: stems and probabilities with different CLIR methods

CLEF topic consists of three parts: `title`, `description` and `narrative`. An example is given below:

    <num> C001
    <title> Architecture in Berlin
    <description> Find documents on architecture in Berlin.
    <narrative> Relevant documents report, in general, on the architectural
features of Berlin or, in particular, on the reconstruction of some
parts of the city after the fall of the Wall.

We used only the `title` and `description` part of the topics and concatenated these simply to form the queries. Table 6 lists some statistics on the test collection[16].

|  | French | English | Italian |
| --- | --- | --- | --- |
| Document source | Le Monde | LA Times | La Stampa |
| # documents | 44,013 | 110,250 | 58,051 |
| # topics | 124 | 122 | 125 |
| # relevant documents | 1189 | 2256 | 1878 |

**Table 6**
Statistics on the test collection

The documents are submitted to the same pre-processing (stemming/lemmatization) procedure as we described in Section 3.1.2. However, for English and French lemmatization, we used the Xelda tools from XRCE[17], which perform morphological normalization slightly differently from the one described in Section 3.1.2. However, since the two lemmatization strategies are based on the same principle (POS-tagging plus inflection removal), the small differences in morphological dictionaries and POS-tagging had no significant influence on retrieval effectiveness.[18]

All runs use a smoothing parameter $\lambda = 0.7$. This value had shown to work well for experiments with several other collections.

### 5.4 Measuring retrieval effectiveness

The effectiveness of retrieval systems can be evaluated by several measures. The basic measures are precision and recall, which cannot be applied directly since they assume clearly separated classes of relevant and non-relevant documents. The most widely accepted measure for evaluating effectiveness of ranked retrieval systems is the average uninterpolated precision, most often referred to as mean average precision (MAP) since the measure is first averaged over relevant documents and then across topics. Other measures, such as precision at a fixed rank, interpolated precision or R-precision, are strongly correlated to the mean average precision, so do not really provide additional information (Tague-Sutcliffe and Blustein, 1995; Voorhees, 1998).

The average uninterpolated precision for a given query and a given system version can be computed as follows: First identify the rank number $n$ of each relevant document in a retrieval run. The corresponding precision at this rank number is defined as the number of relevant documents found in the ranks equal to or higher than the respective rank $r$ divided by $n$. Relevant documents which are not retrieved receive a precision of zero. The average precision for a given query is defined as the average value of the precision $pr$ over all known relevant documents $d_{ij}$ for that query. Finally, the mean av-

---

16 Topics without relevant documents in a sub-collection were discarded.
17 http://www.xrce.xerox.com/competencies/ats/xelda/summary.html
18 We have not been able to substantiate this claim with quantitative figures but did analyze the lemmas that were not found in the translation dictionaries during query translation. We did not find any structural mismatches.

erage precision can be calculated by averaging the average precision over all $M$ queries:

$$\text{MAP} = \frac{1}{M} \sum_{j=1}^{M} \frac{1}{N_j} \sum_{i=1}^{N_j} \text{pr}(d_{ij}) \quad \text{where} \quad \text{pr}(d_{ij}) = \begin{cases} \frac{r_{n_i}}{n_i} & \text{if } d_{ij} \text{ retrieved and } n_i \leq C \\ 0 & \text{in other cases} \end{cases}$$
(14)

Here, $n_i$ denotes the rank of the document $d_{ij}$, which has been retrieved and is relevant for query $j$, $r_{n_i}$ is the number of relevant documents found up to and including rank $n_i$, $N_j$ is the total number of relevant documents of query $j$, $M$ is the total number of queries and $C$ is the cut-off rank ($C$ is 1000 for TREC experiments).

Since we compared many different system versions, which do not always display a large difference in effectiveness, it is desirable to perform significance tests on the results. However, it is well known that parametric tests for data resulting from IR experiments are not very reliable, since the assumptions of these tests (normal or symmetric distribution, homogeneity of variances) are usually not met. We checked the assumptions for an analysis of variance (by fitting a linear model for a within subjects design) and found that indeed the distribution of the residual error was quite skewed, even after transformation of the data. Therefore, we resorted to a non-parametric alternative for the analysis of variance, the Friedman test (Conover, 1980). This test is preferable for the analysis of groups of runs instead of multiple sign-tests or Wilcoxon signed-rank tests, since it provides overall $\alpha$ protection. This means that we first test whether there is any significant difference at all between the runs, before applying multiple comparison tests. Applying just a large number of paired significance tests at the 0.05 significance level without a global test leads very quickly to a high overall $\alpha$. After applying the Friedman test, we ran Fisher's LSD multiple comparison tests (recommended by Hull) to identify equivalence classes of runs (Hull, 1993; Hull, Kantor, and Ng, 1999). An equivalence class is a group of runs which do not differ significantly.

**5.5 Baseline systems**

We decided to have two types of baseline runs. It is standard practice to take a monolingual run as a baseline. This run is based on an IR system using document ranking formula (6). Contrary to runs described in (Kraaij, 2002), we did not use any additional performance enhancing devices, like document length-based priors or fuzzy matching in order to focus on just the basic retrieval model extensions, avoiding interactions.

Systran was used as an additional cross-language baseline, to serve as a reference point for cross-language runs. Notice that the lexical coverage of MT systems varies considerably across language pairs, in particular, the French-English version of Systran is quite good in comparison with other language pairs. We accessed the Web-based version of Systran (December 2002), marketed as "Babelfish", using the Perl utility `babelfish.pm` and converted the Unicode output to the ISO-latin1 character-set to make it compatible with the Xelda-based morphology.

**5.6 Results**

Table 7 lists the results for the different experimental conditions in combination with a translation model pruned with the probability threshold criterion $P > 0.1$ (cf. Section 3.2). For each run, we computed the mean average precision using the standard evaluation tool `trec_eval`. We ran Friedman tests on all the runs based on the Web translation models, because these are the runs we are most interested in; furthermore, one should avoid adding runs that are quite different to a group which is relatively homogeneous, since this would easily lead to a false global significance test. The Friedman test (as measured on the F distribution) proved significant at the $P < 0.05$ level in all cases, so we created equivalence classes using Fisher's LSD method, which are denoted by letters. Letters are assigned in decreasing order of performance; so if a run is member of equivalence class 'a' it is one of the best runs for that task.

The last four rows of the table provide some additional statistics on the query translation process. For both the forward ($P(t|s)$,fw) and the reverse ($P(s|t)$,rev) translation model, we list the percentage of missed translations (% missed)[19] of unique query terms and the average number of translations (# translations) per unique query term.

| run id | EN-FR | FR-EN | EN-IT | IT-EN. |
| --- | --- | --- | --- | --- |
| MONO | 0.4233 | 0.4705 | 0.4542 | 0.4705 |
| MT | 0.3478 | 0.4043 | 0.3060 | 0.3249 |
| QT | a:**0.3760** | a:**0.4126** | a,b:0.3298 | a:**0.3526** |
| DT | a:0.3677 | a,b:0.4090 | a:**0.3386** | a,b:0.3328 |
| SYN | a:0.3730 | b,c:0.3987 | a,b:0.3114 | b:0.3498 |
| QT-EQ | a:0.3554 | a,b:0.3987 | c,d:0.3035 | b,c:0.3299 |
| QT-BM | a:0.3463 | c,d:0.3769 | b,c:0.3213 | b:0.3221 |
| NAIVE | b:0.3303 | d:0.3596 | d:0.2881 | c:0.3183 |
| % missed fw | 9.6 | 13.54 | 16.79 | 9.17 |
| % missed rev | 9.08 | 14.04 | 15.48 | 11.31 |
| # translations fw | 1.65 | 1.66 | 1.86 | 2.13 |
| # translations rev | 22.72 | 29.6 | 12.00 | 22.95 |

**Table 7**
mean average precision and translation statistics ($P > 0.1$)

Table 8 lists the results for the same experimental conditions, but this time the translation models were pruned by taking the $n$ best translation relations according to an entropy criterion, where $n$=100.000 (100K).

| run id | EN-FR | FR-EN | EN-IT | IT-EN. |
| --- | --- | --- | --- | --- |
| MONO | 0.4233 | 0.4705 | 0.4542 | 0.4705 |
| MT | 0.3478 | 0.4043 | 0.3060 | 0.3249 |
| DT | a:**0.3909** | a:0.4073 | a:**0.3728** | a:0.3547 |
| QT | a,b:0.3878 | a:**0.4194** | a:0.3519 | a:**0.3678** |
| QT-BM | b:0.3436 | b:0.3702 | b:0.3236 | b:0.3124 |
| SYN | c:0.3270 | b:0.3643 | b:0.2958 | c:0.2808 |
| QT-EQ | c:0.3102 | b:0.3725 | c:0.2602 | c:0.2595 |
| NAIVE | d:0.2257 | c:0.2329 | d:0.2281 | d:0.2021 |
| % missed fw | 11.04 | 14.65 | 16.06 | 9.36 |
| % missed rev | 10.39 | 16.81 | 15.76 | 10.53 |
| # translations fw | 7.04 | 7.00 | 6.36 | 7.23 |
| # translations rev | 10.51 | 12.34 | 13.32 | 17.20 |

**Table 8**
Mean Average Precision and translation statistics (best 100K parameters)

Several other similar pruning methods have also been tested on the CLEF-2000 subset of the data, e.g. "P>0.01", "P>0.05", "1M parameters", "10K parameters", etc. However, the two cases shown in tables 7 and 8 represent the best of the two families of pruning techniques. Our goal was not to do extensive parameter tuning in order to find the best performing combination of models, but rather to detect some broad characteristics of the pruning methods and their interactions with the retrieval model.

---

19 This figure includes proper nouns.

Since the pruned forward and reverse translation models yield different translation relations (cf. Table 5), we hypothesized that it might be effective to combine both. Instead of combining the translation probabilities directly we chose to combine the results of the QT and DT by interpolation of the document scores. Results for combinations based on the 100K models are listed in Table 9. Indeed, for all the language pairs, the combination run improves upon each of its component runs. This means that each component run can compensate for lexical holes in the companion translation model.

| run id | EN-FR | | FR-EN | | EN-IT | | IT-EN | |
|---|---|---|---|---|---|---|---|---|
| MONO | 0.4233 | | 0.4705 | | 0.4542 | | 0.4705 | |
| MT | 0.3478 | (82%) | 0.4043 | (86%) | 0.3060 | (67%) | 0.3249 | (69%) |
| DT+QT | 0.4042 | (96%) | 0.4273 | (87%) | 0.3837 | (84%) | 0.3785 | (80%) |

**Table 9**
Mean Average Precision of combination run, compared to baselines

### 5.7 Discussion

**5.7.1 Web-based CLIR vs. MT-based CLIR** Our first observation when examining the data is that the runs based on translation models are comparable to or better than the Systran run. Sign tests showed that there was no significant difference between the MT and QT runs for EN-FR and FR-EN language pairs. The QT runs were significantly better at the P=0.01 level for the IT-EN and EN-IT language pairs.

This is a very significant result, particularly since the performance of CLIR with Systran has often been among the best in the previous CLIR experiments in TREC and CLEF. These results show that the Web-based translation models are effective means for CLIR tasks. The better results obtained with the Web-based TM confirm our intuition stated in the introduction that there are better tools for query translation in CLIR than off-the-shelf commercial MT systems.

Comparing to the monolingual runs, the best CLIR performance with Web-TM varies from 74.1% to 93.7% ( 80% to 96% for the combined QT+DT models) of the monolingual run. This is within the typical range of CLIR performance. More generally, this research successfully demonstrates the enormous potential of parallel Web pages and Web-based MT.

We cannot really compare performance across target languages, since the relevant documents are not distributed in a balanced way: some queries do not have any relevant document in some languages. This partly explains why the monolingual IT-IT run is much higher than the monolingual French and English runs. We can, however, compare methods within a given language pair.

**5.7.2 Comparison of query model translation (QT), document model translation (DT) and translations modeled as synonyms (SYN)** Our second question in Section 5.1 concerned the relative effectiveness of the QT and DT models. The experimental results show that there is no clear winner; differences are small and not significant. There seems to be some correlation with translation direction, however: the QT models perform better than DT on the X-EN pairs and the DT models perform better on the EN-X pairs. This might indicate that the $P(e|f)$ and $P(e|i)$ translation models are more reliable than their reverse counterparts. A possible explanation for this could be that the average English

sentence is shorter than a French and Italian sentence. The average number of tokens per sentence is respectively 6.6/6.9 and 5.9/6.9 for EN/FR and EN/IT corpora. This may lead to more reliable estimates for $P(e|f)$ and $P(e|i)$ than the reverse. However, further investigation is needed to confirm this, since differences in morphology could also contribute to the observed effect. Still the fact that QT models perform just a good as DT models in combination with translation models is a new result.

We also compared our QT and DT to the synonym-based approach (Pirkola, 1998). Both the QT and DT model were significantly more effective than the synonym based model (SYN). The latter seems to work well when the number of translations is relatively small, but cannot effectively handle the large number of (pseudo)-translations as produced by our 100K translation models. The synonym based model performs usually better than the models based on query translation with uniform probabilities, but differences are not significant in most cases.

**5.7.3 Query expansion effect** In the introduction we argued that using just one translation (as MT does) is probably a suboptimal strategy for CLIR, since there is usually more than one good translation for a term. Looking at probabilistic dictionaries, we have also seen that the distinction between a translation and a closely related term cannot really be made on the basis of some thresholding criterion. Since it is well known in IR that adding closely related terms can improve retrieval effectiveness, we hypothesized that adding more than one translation would also help. The experimental results confirm this effect. In all but one case (EN-FR, $P > 0.1$) using all translations (QT) yielded significantly better performance than choosing just the most probable translation (QT-BM). For the $P > 0.1$ models, the average number of translations in the forward direction is only 1.65, so the potential for a query expansion effect is limited, which could explain the non-significant difference for the EN-FR case.

Unfortunately, we cannot say whether the significant improvement in effectiveness is mainly due to the fact that the probability of giving at least one good translation (which is probably the most important factor for retrieval effectiveness (Kraaij, 2002; McNamee and Mayfield, 2002)) is higher for QT or indeed to the query expansion effect. A simulation experiment is needed to quantify the relative contributions. Still, it is of great practical importance that more (weighted) translations can enhance retrieval effectiveness significantly.

**5.7.4 Pruning & weighting** A related issue is the question of whether it is more important to prune translations or to weight them. Grefenstette (cf. Section 1) originally pointed out the importance of pruning and weighting translations for dictionary-based CLIR. Pruning was seen as a means of removing unwanted senses in a dictionary-based CLIR application. Our experiments confirm the importance of pruning and weighting, but in a slightly different manner. In a CLIR approach based on a Web translation model, the essential function of pruning is to remove spurious translations. Polluted translation models will result in a very poor retrieval effectiveness. As far as sense disambiguation is concerned, we believe that our CLIR models can handle sense ambiguity quite well. Our best performing runs, based on the 100K models, have on average seven translations per term! Too much pruning (e.g. best match) is sub-optimal. However, the more translation alternatives we add, the more important their relative weighting becomes.

We have compared weighted translations (QT) with uniform translation probabilities (QT-EQ). In each of the eight comparisons (four language pairs, two pruning techniques), weighting results in a improved retrieval effectiveness. The difference is significant in six cases. Differences are not significant for the $P < 0.1$ EN-FR and FR-EN translation models. We think this is due to the small number of translations; a uniform translation probability will not differ radically from the estimated translation probabilities.

The importance of weighting is most evident when the 100K translation models are used. These models yield seven translations on average for each term. The CLIR models based on weighted translations are able to exploit the additional information and show improved effectiveness w.r.t. the $P < 0.1$ models. The performance of unweighted CLIR models (QT-EQ and SYN) is seriously impaired by the higher number of translations.

The comparison of the naive dictionary-like replacement method, which does not involve any normalization for the number of translations per term (NAIVE), with QT-EQ shows that normalization (i.e. a minimal probabilistic embedding) is essential. The NAIVE runs have the lowest effectiveness of all variant systems (with significant differences). Interestingly, it seems better to select just the one most probable translation than taking all translations unweighted.

**5.7.5 Robustness**  We pointed out in the previous section that the weighted models are more robust, in the sense that they can handle a large number of translations. We found however that the query model translation method (QT) and the document model translation method (DT) display a considerable difference in robustness to noisy translations. Initially we expected that the DT method (where the matching takes place in the source language) would yield the best results, since this model has previously proven to be successful for several quite different language pairs, e.g. European languages, Chinese and Arabic using parallel corpora or dictionaries as translation devices (McNamee and Mayfield, 2001; Xu, Weischedel, and Nguyen, 2001; Hiemstra et al., 2001).

However, our initial DT runs obtained extremely poor results. We discovered that this was largely due to noisy translations from the translation model (pruned by the $P < 0.1$ or 100K method) which is based on Web data. There are many terms in the target language, which occur very rarely in the parallel Web corpus. The translation probabilities for these terms (based on the most probable alignments) are therefore unreliable. Often these rare terms (and non-words like xc64) are aligned with more common terms in the other language and are not pruned by the default pruning criteria ($P > 0.1$ or best 100K parameters), since they have high translation probabilities. This especially poses a problem for the DT model, since it includes a summation over all terms in the target language that occur in the document and have a non-zero translation probability. We devised a supplementary pruning criterion to remove these noisy translations, discarding all translations for which the source term has a marginal probability in the translation model which is below a particular value (typically $10^{-6} - 10^{-5}$). Later we discovered that a simple pruning method was even more effective: discard all translations where either the source or target term contains a digit. The results in Tables 7 and 8 are based on the latter additional pruning criterion. The QT approach is less sensitive to noisy translations arising from rare terms in the target language, because it is easy to remove these translations using a probability threshold. We deduce that extra care therefore has to be taken to prune translation models for the document model translation approach to CLIR.

## 6 Conclusions

Statistical translation models require large parallel corpora and unfortunately, only a few manually constructed ones are available. In this paper, we have explored the possibility of automatically mining the Web for parallel texts in order to construct such corpora. Translation models are then trained on these corpora. We subsequently examined different ways to embed the resulting translation models in a cross-language information retrieval system.

To mine parallel Web pages, we constructed a mining system called PTMiner. This system employs a series of heuristics to locate candidate parallel pages and determine if they are indeed parallel. We have successfully used PTMiner to collect several corpora for different language pairs: English-French, English-Italian, English-German, English-Dutch and English-Chinese. The language-independent characteristics of PTMiner allowed us to adapt it quite easily to different language pairs.

The heuristics used in the mining process seem to be effective. Although it cannot collect all pairs of parallel pages, our preliminary evaluation shows that the system's precision is quite good. The recall ratio is less important in this context because of the abundance of parallel pages on the Web.

The mining results - parallel corpora - are subsequently used to train statistical translation models, which are exploited in a CLIR system. The major advantage of this approach is that it can be fully automated, avoiding the tedious work of manual collection of parallel corpora. On the other hand, compared to manually prepared parallel corpora, our mining results contain more noise (i.e. non-parallel pages). For a general translation task this may be problematic; but for CLIR however, the noise contained in the corpora is less dramatic. In fact, IR is strongly error-tolerant. A small proportion of incorrect translation words can be admitted without a major impact on global effectiveness. Our experiments showed that a CLIR approach based on the mined Web corpora can in fact outperform a good MT system (Systran). This confirms our initial hypothesis that noisy parallel corpora can be very useful for applications such as CLIR. Our demonstration that the Web can indeed be used as a large parallel corpus for tasks such as CLIR is the main contribution of this paper.

Most previous work on CLIR has separated the translation stage from the retrieval stage, i.e. query translation is considered as a preprocessing step for monolingual IR. In this paper, we have integrated translation and retrieval within the same framework. The advantage of this integration is that we do not need to obtain the optimal translation of a source query, and then an optimal retrieval result given a query translation, but instead aim for the optimal global effect. The comparisons between our approach and simulated external approaches clearly show that an integrated approach performs better.

We also compared two ways of embedding translation models within a CLIR system: i) translating the source query model into the target (document) language; ii) translating the document model into the source language. [20] Both embedding methods produced very good results Compared to our reference run with Systran. However, it is still too early to assert which embedding method is superior. However, we did observe a significant difference in robustness between them: the document-model translation method is much more sensitive to spurious translations, since the model incorporates all source terms that have a non-zero translation probability into a query term. We devised two supplementary pruning techniques that effectively removed the noisy terms: either by removing terms containing digits, or by removing translations based on source terms with a low marginal probability. (This latter is perhaps more principled.)

On the use of statistical translation models for CLIR, we have demonstrated that this naturally produces a desired query expansion effect, resulting in finding more related documents. In our experimental evaluation, we saw that it is usually better to include more than one translation, and to weigh these translations according to the translation probabilities, rather than using the resulting translation model as a bilingual lexicon for external translation. This effect partly accounts for the success of our approach in comparison with an MT-based approach, which only retains one translation per sense. However, this technique should not be exaggerated; otherwise, too much noise will be introduced. To avoid this, it is important to incorporate pruning.

---

[20] Another way which interprets multiple translations as synonyms is a special case of the latter.

We investigated several ways to prune translation models. The best results were obtained with a pruning method based on the top 100K parameters of the translation model. These models produced more than seven translations per word on average, demonstrating the capability of the CLIR model to handle translation ambiguity and exploit co-occurrence information from the parallel corpus for query expansion.

There are several ways in which our approach can be improved. First, regarding PTMiner, more or better heuristics could be integrated in the mining algorithm. As we mentioned, some parallel Web sites are not always organized in the way we would expect. This is particularly the case for non-European languages such as Chinese and Japanese. Hence, one of the questions we wish to investigate is how to extend the coverage of PTMiner to more parallel Web pages. One possible improvement would be to integrate a component that "learns" the organization patterns of a Web site, assuming, of course, that a Web site is organized in a consistent way. Preliminary tests have shown that this is possible to some extent: we can recognize dynamically that the parallel pages on `www.operationid.com` are at `www.operationcarte.com` or that the file `index1.html` corresponds to `index2.html`. Such criteria complement the ones currently employed in PTMiner.

In its current form, PTMiner scans parallel candidates according to similarities in file names. This step does not exploit the hyperlinks between the pages; whereas we know that two pages which are referenced at comparable structural positions in two parallel pages have a very high chance of themselves being parallel. This could well improve the quality of PTMiner.

When the mining results are not fully parallel, it would be interesting to attempt to clean them in order to obtain a higher-quality training material. One possible approach is to use sentence alignment as an additional filter, as we mentioned earlier. This approach has been applied successfully to our English-Chinese Web corpus. The cleaned corpus results in both higher translation accuracy and higher CLIR effectiveness. However, this approach has still to be tested for the European languages.

In this study, we hypothesized that IBM *Model 1* is appropriate for CLIR, primarily because word order is not important for IR. Although it is true that word order is not important in current IR approaches, it is definitely important to consider context words during the translation. For example, when deciding how to translate the French word "tableau" (which may refer to a painting, a blackboard, a table [of data], etc.), if we observe "artistique" (artistic) next to it, then it is pretty certain that "tableau" means a painting. A more sophisticated translation model than IBM *Model 1* could produce a better selection of translation words.

We also rely solely on word translation in our approach, although it is well known that this simplistic approach cannot correctly translate compound terms such as "pomme de terre" (potato) and "cul de sac" (no exit). Incorporating the translation of compound terms in a translation model should result in additional improvements for CLIR. Our preliminary experiments (Nie and Dufort, 2002) on integrating the translation of compounds certainly showed this, with improvement of up to 70% over a word-based approach. This direction warrants further investigation.

Finally, all our efforts thus far to mine parallel Web pages have involved English. How can we deal with CLIR between, say Chinese and German, for which there are few parallel Web sites? One possible solution would be to use English as a pivot language, even though such a two-step translation would certainly reduce accuracy and introduce more noise. Nevertheless, several authors have shown that a pivot approach can still produce effective retrieval and can at least complement a dictionary-based approach (Franz, McCarley, and Ward, 2000; Gollins and Sanderson, 2001; Lehtokangas and Airio, 2002).

## 7 Acknowledgements

This work was partly funded by a research grant from the Dutch Telematics Institute: DRUID project. We would like to thank Xerox Research Center Europe (XRCE) for making their Xelda toolkit available to us. We would also like to thank George Foster for making his statistical MT toolkit available and for many interesting discussions. Special thanks are due to Jiang Chen who contributed in building PTMiner. Finally, we want to thank Elliott Macklovitch and the two anonymous reviewers for their constructive comments and careful review. Part of this work was carried out while the first author was visiting the RALI laboratory at Université de Montréal.